# Long Exposure Localization in Darkness Using Consumer Cameras

Michael J. Milford, *Member, IEEE*, Ian Turner, Peter Corke, *Fellow, IEEE*

*Abstract*—In this paper we evaluate performance of the SeqSLAM algorithm for passive vision-based localization in very dark environments with low-cost cameras that result in massively blurred images. We evaluate the effect of motion blur from exposure times up to 10,000 ms from a moving car, and the performance of localization in day time from routes learned at night in two different environments. Finally we perform a statistical analysis that compares the baseline performance of matching unprocessed grayscale images to using patch normalization and local neighborhood normalization – the two key SeqSLAM components. Our results and analysis show for the first time *why* the SeqSLAM algorithm is effective, and demonstrate the potential for cheap camera-based localization systems that function despite extreme appearance change.

## I. INTRODUCTION

One of the currently accepted norms in robotics and computer vision research is that visual sensors become ineffective in poor lighting. In low lighting situations exposure duration or camera gain must generally be increased to obtain an image with an appropriate level of brightness and contrast. Increasing either of these parameters has negative side effects; increasing the exposure duration leads to blurry images if the camera is moving, while increasing the gain leads to a noisier image. These side effects are both potentially catastrophic for many types of vision processing techniques, especially those that rely on the now standard gradient-based feature detection algorithms such as Scale-Invariant Feature Transforms (SIFT) [1] and Speeded Up Robust Features (SURF) [2]. A range of solutions have been proposed including high dynamic range techniques, high sensitivity and thermal cameras, active lighting/strobing of the environment, or simply using alternative sensors such as laser rangefinders. However, each of these solutions has one or more significant disadvantages including, but not limited to: prohibitive cost, intrusiveness, ineffectiveness on fast moving platforms, power consumption and bulkiness.

In this paper, we propose sacrificing both image sharpness and quality by maximizing the camera's exposure duration and gain, in order to obtain well-exposed images using relatively cheap consumer hardware (Fig. 1). Using the SeqSLAM localization algorithm [3], we conduct a range of experimental studies that show, perhaps surprisingly, that place recognition along a route is largely invariant to motion blur and that places visited during the day can be visually recognized at night in almost pitch black conditions. We make the following contributions:

- a study demonstrating that low resolution image matching performance is largely invariant to motion blur, even if the images being matched differ in exposure duration by an order of magnitude,

- experimental evaluation of localization in two different night time environments using two consumer cameras, including an unlit environment two orders of magnitude darker than the previous benchmark result [3], and

- a statistical analysis of image matching performance on a day-night dataset, showing firstly that low resolution grayscale images are by themselves uninformative, and secondly how patch normalization and local neighborhood normalization turn these uninformative images into highly spatially salient information. This analysis provides for the first time an explanation of why the SeqSLAM algorithm works.

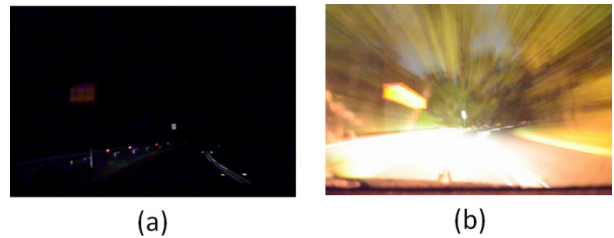

Fig. 1 – By maximizing a camera's exposure duration and gain, a correctly exposed image can be obtained even in a pitch black environment, at the cost of both image sharpness and quality. In this paper we show how these blurry, noisy images can be processed to provide highly salient localization information.

The work in this paper extends recent work [3] that introduced the SeqSLAM algorithm and set the benchmark for passive, conventional camera-based localization across day-night cycles. Successful vision-based localization was achieved using relatively sharp imagery obtained on well illuminated main roads at night. In this work we revisit and successfully localize along poorly lit suburban backstreets on which the previous approach [3] failed due to the inability to sufficiently expose images. We then use a consumer camera with a larger sensor and even longer exposure durations to demonstrate localization in a much darker unlit environment. We also provide analysis of why the SeqSLAM algorithm is effective.

M.J. Milford and P. Corke are with the School of Electrical Engineering and Computer Science at the Queensland University of Technology, Brisbane, Australia, michael.milford@qut.edu.au. I. Turner is with the School of Mathematical Sciences at the Queensland University of Technology. This work was supported by an Australian Research Council Fellowship DE120100995 to MM.

The paper proceeds as follows. Section II provides some background on vision-based localization and mapping techniques relevant to the presented approach. In Section III we briefly describe the SeqSLAM algorithm. Section IV presents the experimental setup and the two testing environments. Results including quantitative image match performance and sample frame matches are provided in Section V. In Section VI we discuss the significance of the results and outline areas for future work, before the paper concludes in Section VII.

## II. Background

We skip over a general review of vision-based mapping systems and instead touch on related research in three areas: use of image sequences rather than single frames to perform localization, low resolution visual navigation, and illumination invariant vision-processing techniques. Sequence matching has been performed in a navigation context in [4], where sequences of images were compared on the basis of 128D SIFT descriptor vectors. Excessive feature ambiguity was managed by additional algorithms to achieve reliable loop closure. The use of image sequences has also been used in biologically-based navigation systems [5, 6], to map environments such as a city suburb using 2D pixel intensity profiles [6]. Low resolution images have been used to achieve navigation in applications [7-9] such as autonomous car driving on roads using 30×32 pixel images. While most of these approaches are not reliant on feature detection, to the best of our knowledge no-one has investigated their performance on long exposure images.

Most of the work on making feature detection more robust to illumination change has focused on technological solutions to obtain sharp images. High dynamic range approaches to vision-based localization improve the information content of an image [10]. However, such techniques still require enough light – otherwise sensor gains and exposure durations must be increased, resulting in noise and motion blur if the camera is moving, which is inevitable in any navigation scenario. Vision-based techniques have also been coupled with range sensors using multisensory fusion. These approaches require that the same features are detectable by both vision and range sensors [11, 12]. Once again, if the environment is dark enough to require longer exposures, images blur and common feature detection becomes difficult. Although there are techniques for recovering camera motion from blurred images [13, 14], these are only applicable for either pitch-roll-yaw type movement, or camera exposures during translational motion over relatively short distances, much shorter than those presented here.

If we wish to start with crisp and bright images from low light environments and a moving platform the only options are to apply light to the scene or use a highly sensitive camera. Active lighting negates the passive advantage of a camera over other sensors such as laser range finders, and is often not practical for reasons such as energy consumption. More sensitive cameras require a larger and/or more sensitive image sensor and better lens, but these are prohibitively expensive. Other sensors such as laser range finders, are expensive, active, and provide a relatively sparse scan of the world, and using thermal images across day and night cycles is challenging [15]. In this work we explore an alternative way forward; we accept the poor image quality and develop robust algorithms that accommodate them.

## III. SeqSLAM

In this section we describe the primary components of the SeqSLAM algorithm. In brief, each new image is compared to all previous images to produce an image difference vector. These vectors are accumulated to form an image difference matrix, which is searched for sequences containing low average difference scores, corresponding to sequence match candidates.

### A. Image Comparison

Although SeqSLAM performs loop closure by matching coherent *sequences* of images rather than individual images, it still requires an individual image comparison method. In this work, we used a Sum of Absolute Differences (SAD) calculation on resolution reduced, patch-normalized images (Fig. 2) to produce an image difference score *d*:

$$d(A,B) = \frac{1}{R_x R_y} \sum_{x=1}^{R_x} \sum_{y=1}^{R_y} \left| \mathbf{A}_{x,y} - \mathbf{B}_{x,y} \right| \quad (1)$$

where $R_x$ and $R_y$ are the dimensions of the resolution reduced image, and **A** and **B** are matrices containing the patch-normalized grayscale pixel intensity values for the two images being compared. Patch normalization is defined by:

$$A'_{x,y} = \frac{A_{x,y} - \bar{A}}{\sigma_D} \quad (2)$$

where

$$\bar{A} = \frac{1}{|W|} \sum_{i,j \in \mathbb{W}(x,y)} A_{i,j}, \quad \sigma_D = \frac{1}{|W|} \sum_{i,j \in \mathbb{W}(x,y)} (A_{i,j} - \bar{A})^2 \quad (3)$$

and $\mathbb{W}(x,y)$ is an $n_p \times n_p$ window centered on $(x, y)$. Patch normalization is applied at discrete intervals of distance $n_p$ in each direction.

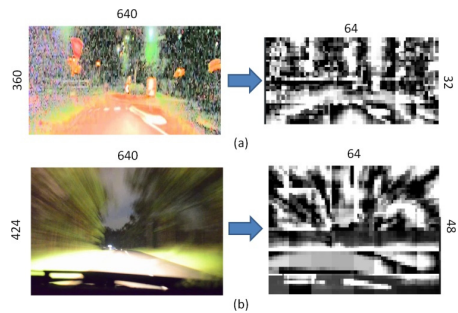

Fig. 2: Original images and low resolution patch-normalized versions used by SeqSLAM for the a) backstreets and b) Mt Cootha datasets.

### B. Template Learning

As the algorithm processes images, it stores resolution reduced, patch-normalized images as visual templates. For all experiments performed in this paper, templates were learned for every single frame of the dataset being processed. Each new frame is compared to all the existing visual templates using the image comparison method described in

Section IIIA. This comparison produces a vector of image differences:

$$\mathbf{D}^i = [d(T_1, T_i), d(T_2, T_i), \ldots d(T_{i-1}, T_i)] \quad (4)$$

The length of the image difference vector grows linearly as more frames are processed.

### C. Local Neighborhood Normalization

Due to large variations in overall scene lighting and composition, the image difference vector can have systematic biases. For example, a day-time image might match more closely to all other day-time visual templates than any of the night-time visual templates due to an overall difference in brightness. To remove this bias, we apply a local contrast enhancement to each element $\mathbf{D}^i_k$ in the image difference vector to produce a new vector:

$$\hat{D}^i_k = \frac{D^i_k - \overline{D}}{\sigma_D} \quad (5)$$

where

$$\overline{D} = \frac{1}{2N+1} \sum_{j=-N}^{N} D^i_k, \quad \sigma_D = \frac{1}{2N} \sum_{j=-N}^{N} (D^i_k - \overline{D})^2 \quad (6)$$

are the local mean and standard deviation in a range of ±$N$ templates around template difference $\mathbf{D}^i_k$. The resultant image difference vector now has templates within every local section of route that strongly match (i.e. have a low difference score) the current image. The analysis in Section V.E shows how local neighborhood normalization significant improves the quality of the individual image matches.

### D. Localized Sequence Recognition

Over time, the contrast enhanced image difference vectors for the $n$ most recent frames form an image difference matrix:

$$\mathbf{M} = \left[ \hat{\mathbf{D}}^{c-n+1}, \hat{\mathbf{D}}^{c-n+2}, \ldots \hat{\mathbf{D}}^c \right] \quad (7)$$

where $c$ is the index of the current frame. Every image difference vector is padded out to the length of the most recent (and largest) vector. This difference matrix can now be searched to find spatially coherent sequences of templates that match corresponding images in the $n$ most recent images.

We perform a lightweight version of the Dynamic Time Warping algorithm (DTW) of Sakoe and Chiba [16]. We apply a constraint on the range of possible slopes, but do not utilize boundary conditions or monotonically increasing constraints. Searches are performed starting at every element in the oldest image difference vector $\hat{\mathbf{D}}^{c-n+1}$. The slope constraint relates to the maximum variation in velocity on repeated traverses of a route (values given in Table II). Consequently, multiple searches are performed from each element in $\hat{\mathbf{D}}^{c-n+1}$ for different slopes within the allowable slope range.

Each search results in a difference score $S(i, m)$, which represents the average image difference over all image pairs between the two image sequences starting at location $I$ in the oldest difference vector and with a slope of $m$. The best matching sequence is determined by:

$$\{s^*, m^*\} = \arg\min_{\substack{1 \leq i \leq p \\ X \leq m \leq Y}} S(i, m) \quad (8)$$

If the minimum score is *below* a threshold $s_m$, then a sequence is deemed to be a match.

## IV. EXPERIMENTAL SETUP

In this section we describe the A) cameras used, B) testing environments, C) studies conducted, D) ground truth measures, E) image pre-processing and F) SeqSLAM parameter values.

### A. Cameras

Two types of cameras were mounted on a car dashboard facing forwards through the windshield. For the backstreets dataset, a Logitech C910 webcam was used, a 100 USD webcam with a 20 mm$^2$ sensor. For the Mt Cootha dataset, a Nikon D5100 camera equipped with 18-55 mm kit lens was used (set at 18 mm) alongside the C910. The D5100 has an APS-C size sensor measuring 368 mm$^2$, approximately 18 times larger in area than the webcam sensor. The camera retails for approximately 650 USD. Both cameras are significantly cheaper than many of the standard industrial cameras widely used in robotics, which are typically 1000-2500 USD. To achieve a 630 ms exposure duration during the day on the larger sensor, we used a 9 F-stop neutral density (ND) filter, which reduced the incoming light intensity by a factor of 512. The 630 ms exposures were captured at a frame rate of 1 frame per second (rather than at the theoretically maximum rate of 1.6 frames per second) due to a slight storage and shutter delay between image captures. Consequently, the camera shutter was closed for 370 ms of every second, meaning only two thirds of the route was actually captured by the camera.

### B. Datasets

A total of five datasets were obtained from two different environments, as shown in Table I. All datasets are available at https://wiki.qut.edu.au/display/cyphy/Michael+Milford+Datasets+and+Downloads. For safety reasons, all the night-time runs were obtained with headlights on. However, the headlights did not illuminate the top two thirds of the image, and overexposed the road surface, providing no useful information to the algorithm. Although not presented here, we have image region analysis results that show the bottom third of the image provided no localization information.

TABLE I  DATASETS

| Data-set | Environ-ment | Time of day | Cam-era | # of Frames | Mean frame spacing | Exposure mode | ISO Rating |
|---|---|---|---|---|---|---|---|
| 1 | Backstreets | Night | C910 | 928 | 1.9 m | 200 ms | Auto |
| 2 | Backstreets | Day | C910 | 977 | 1.8 m | Auto (< 67 ms) | Auto |
| 3 | Mt Cootha | Night | D5100 | 740 | 12.8 m | 630 ms | 25600 |
| 4 | Mt Cootha | Day | D5100 | 724 | 13.1 m | 630 ms | 320 + filter |
| 5 | Mt Cootha | Day | C910 | 1948 | 4.9 m | Auto (< 67 ms) | Auto |

*1) Backstreets Environment*

The backstreets environment consisted of a 1.8 km route along a network of suburban backstreets with sparse street lighting and little other ambient light. Speeds varied between 0 and 45 km/hr.

*2) Mt Cootha Environment*

The Mt Cootha environment consisted of a 9.5 km mountain road loop with mostly no street lighting (Fig. 1). Speeds varied between 0 and 50 km/hr. A range of weather conditions including heavy fog were encountered along parts of the route. Headlights from opposing traffic also overexposed the images at several locations along the route.

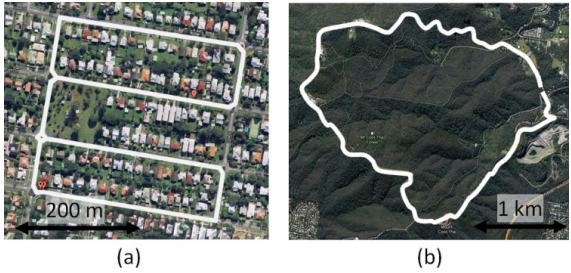

Fig. 3: Aerial photo of the (a) suburban backstreets and (b) Mt Cootha environments. The routes taken are shown by thick white lines. Copyright DigitalGlobe, GeoEye, Getmapping plc, The GeoInformation Group, USDA Farm Service Agency, Infoterra Ltd & Bluesky, Map data ©2012 Google.

*C. Studies*

We conducted four experimental studies using the two datasets. The first study evaluated the effect of motion blur on general day-time localization performance, while the other three studies involved *localizing during the day using visual templates learnt at night*. Although we do not present results here, localization performance at night using visual templates learnt during the day (the inverse situation) was similar.

*1) Variable Motion Blur*

To create arbitrary motion blur we used a moving average temporal blur. For example, to create simulated 10000 ms exposures, we combined a moving window of 150 frames from the original 15 frames per second video that dataset 5 was extracted from. Because a moving average temporal blur provides a smoother image signal than actual discrete long exposures from a real camera, we validate the simulated exposure trials with real long exposure trials. Each of the variable exposure datasets was matched back to dataset 4, the fixed long exposure day-time dataset.

*2) Variable Short and Fixed Medium Duration Exposures*

The second study consisted of running the algorithm on the night run of the backstreets dataset and then the day run of the backstreets dataset, using the C910 webcam for both. The webcam was set to 200 ms exposure durations for the night run and set to auto-expose (maximum exposure duration 67 ms) during the day-time dataset. This study tested whether a low cost webcam at maximum gain and exposure duration would make navigation feasible on dark suburban backstreets.

*3) Fixed Long Duration Exposures*

The third study involved running the algorithm on the night run of the mountain road dataset and then the day run of the mountain road dataset, using the D5100 camera set to 630 ms exposures for both runs. This study tested whether a larger sensor size would enable localization on a road with no street lighting and minimal light from the night sky due to heavy vegetation.

*4) Variable Short and Fixed Long Duration Exposures*

The final study involved running the algorithm on the night run of the mountain road dataset using the D5100 630 ms exposure images and then on the day run of the mountain dataset using the C910 webcam set to auto-expose (maximum exposure duration 67 ms). This scenario tested whether localization was robust to more than an order of magnitude variation in exposure duration, an outcome which would facilitate implementation.

*D. Ground Truth*

GPS was not reliably available throughout either of the two environments. Instead, ground truth frame correspondences were obtained by parsing each video and manually assigning frame correspondences at regular intervals. Linear interpolation provided the ground truth correspondences for in-between frames. Corresponding frames were (manually) identified using distinctive environmental features. The ground truth can be considered to be accurate to half a frame interval. Since a relatively small field of view forward facing camera was used, the frames from forward and backward traverses of the backstreets dataset were tagged as different locations.

Localization errors were calculated by measuring the ground truth distance between the frames at the centre of each matched pair of sequences. The metric error is calculated by multiplying the frame error by the average distance between frames for the dataset.

*E. Image Pre-Processing*

A single rectangular crop was performed on the videos from datasets 1, 2 and 5 to achieve an approximately corresponding field of view. This step was necessary due to the camera placement changing between datasets and also because the C910 and D5100 have different fields of view. No lens distortion correction or other transformations were performed, leaving some inconsistent distortion between the two cameras. 8 bit RGB pixel values $(R,G,B)$ were converted to 8 bit grayscale pixel values $I$ using ITU Rec 709:

$$I = 0.2989R + 0.5870G + 0.1140B \qquad (9)$$

*F. Parameters*

Table II provides the values of the critical parameters used in these experiments:

TABLE II  PARAMETER LIST

| Parameter | Value | Description |
|---|---|---|
| $R_x, R_y$ | 64, 32 | Backstreets environment |
| $R_x, R_y$ | 64, 48 | Mt Cootha environment |
| $R_{window}$ | 10 templates | Local template neighborhood range |
| $n$ | 50 frames / 655 m | Study 1 |
| $n$ | 100 frames / 190 m | Study 2 |
| $n$ | 20 frames / 262 m | Study 3 |
| $n$ | 50 frames / 245 m | Study 4 |
| $V_{min}$ | $0.84V_{av}$ | Minimum sequence speed ratio |
| $V_{max}$ | $1.19V_{av}$ | Maximum sequence speed ratio |
| $V_{step}$ | $0.04V_{av}$ | Sequence ratio step-size |
| $P$ | 8 pixels | Patch normalization patch side length |

The sequence matching length for studies 2-4 was chosen such that the metric distance represented by a sequence at maximum velocity would be similar (a longer sequence was used for study 1 because of the extreme exposure durations).

Because frame rates were inconsistent between some datasets, the search velocity range was calibrated using the average frame rate ratio between the two datasets.

## V. RESULTS

In this section we first present the results of the variable motion blur study, and then the remaining three studies involving long exposure camera imagery. Qualitative results include image template graphs, while quantitative performance is assessed using recall rates and mean and maximum localization errors. For illustrative purposes we also show sample original and patch-normalized images from sequences which were matched by SeqSLAM. The video accompanying the paper shows sequences of frame matches output by the localization algorithm for studies 3 and 4. For all experiments computation was performed at real-time speed or faster on an Intel Core i5 PC in Matlab and C++.

### A. Variable Motion Blur

Localization is surprisingly robust to moving average temporal blur caused by simulated exposure durations of up to 10000 ms. Fig. 4 shows frame matches overlaid on ground truth for all six exposure durations. All frame match graphs up to 5000 ms were generated using a sequence difference threshold that generated no large false positive errors. Only above 5000 ms do false positives start to occur. Not surprisingly, because we were matching to fixed 630 ms exposure images, performance was best for the simulated 500 ms and 1000 ms exposure durations (Table III). Note that the maximum recall achievable was 93.2% due to the algorithm needing a full sequence length before localization could commence.

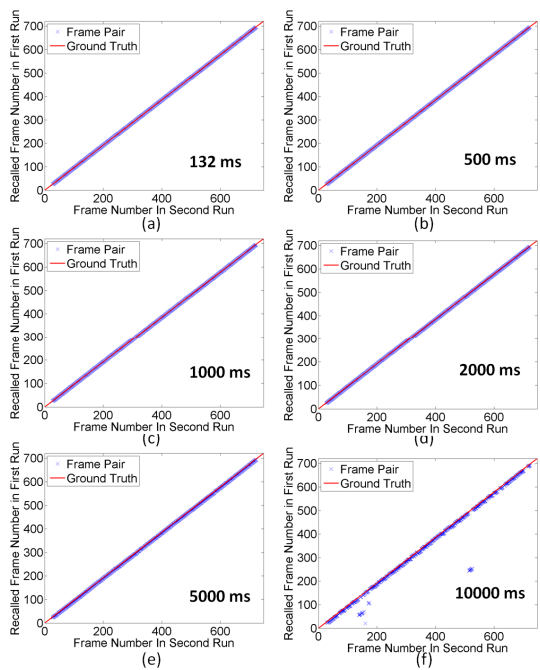

Fig. 4: Matches between the second (varied blur) run and the first fixed exposure run for motion blur corresponding to (a) 132 ms, (b) 500 ms, (c) 1000 ms, (d) 2000 ms, (e) 5000 ms and (f) 10000 ms exposure durations.

The gradually increasing mean and maximum localization errors for 1000 ms and longer durations led us to examine a zoomed in section of the frame matching graphs (Fig. 5). The graph clearly shows there is a lag in the frame matching, which increases as the degree of motion blur increases. Upon consideration, this effect is to be expected, as longer and longer exposure times will create an image which represents a temporal average of images further and further backwards in time.

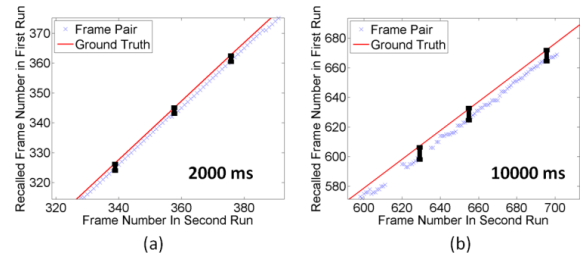

Fig. 5: Zoomed in section of the image match graphs shown in Fig. 4 for the more severe motion blur experiments: (a) 2000 ms (b) 10000ms. The black vertical bars show the offset between the ground truth matches (solid red line) and the reported matches, with the offset increasing as the degree of motion blur increases.

TABLE III  MOTION BLUR RECALL RATES AND LOCALIZATION ERRORS

| EXPOSURE LENGTH | RECALL | MEAN LOCALIZATION ERROR | | MAX LOCALIZATION ERROR | |
|---|---|---|---|---|---|
| | | FRAMES | METERS | FRAMES | METERS |
| 132 ms | 93.2% | 0.44 | 5.8 | 1.38 | 18 |
| 500 ms | 93.2% | 0.376 | 5.0 | 1.35 | 18 |
| 1000 ms | 93.2% | 0.410 | 5.4 | 1.71 | 23 |
| 2000 ms | 93.2% | 0.797 | 10.5 | 2.22 | 29 |
| 5000 ms | 93.2% | 2.46 | 32.4 | 4.27 | 56 |
| 10000 ms | 87.3% | 11.5 | 152 | 252 | 3320 |

### B. Variable Short and Fixed Medium Duration Exposures

Table IV shows the maximum recall rates achieved without any large localization errors for Studies 2 to 4 on the Mt Cootha datasets (see max localization errors). Although the frame errors are broadly similar between all three scenarios, the metric error is much smaller for the backstreets dataset due to the higher frame-rate. Just over half of the locations were reliably matched to within an average of one frame within the backstreets dataset, with a maximum matching error of 3.1 frames. Fig. 6 shows the corresponding frames for a matched day-night sequence. Note the dissimilarity in the images from the matched sequence, both in terms of the original images and the grayscale images.

TABLE IV  RECALL RATES AND LOCALIZATION ERRORS

| Dataset | Recall | Mean Localization Error | | Max Localization Error | |
|---|---|---|---|---|---|
| | | Frames | Meters | Frames | Meters |
| Backstreets | 50.5% | 0.8 | 1.5 | 3.1 | 5.9 |
| Mt Cootha – all long exposures | 76% | 1.1 | 14 | 3.5 | 46 |
| Mt Cootha – mixed exposures | 79% | 0.62 | 8.1 | 1.6 | 21 |

### C. Fixed Long Duration Exposures

76% of locations were correctly recalled to with an average of 1.1 frames for the uniform long exposure Mt Cootha experiment (study 3).

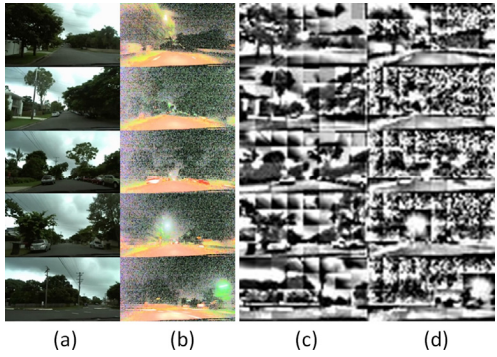

Fig. 6: Corresponding frames for a matching day-night backstreets sequence (datasets 1 and 2). (c-d) The actual patch normalized images used by the comparison algorithm.

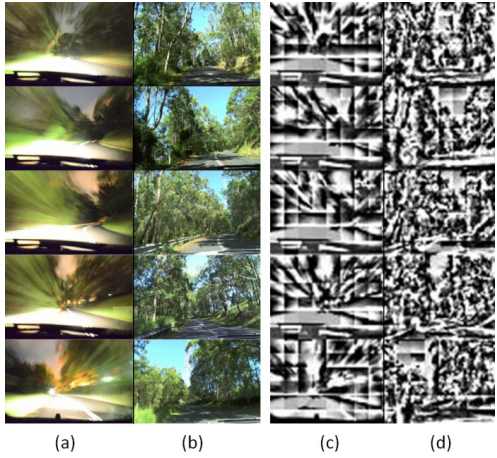

Fig. 7: Corresponding frames for a matched day-night image sequence between the (a) long exposure D5100 images and the (b) short exposure C910 webcam images (datasets 3 and 5). (c-d) The actual patch normalized images used by the comparison algorithm.

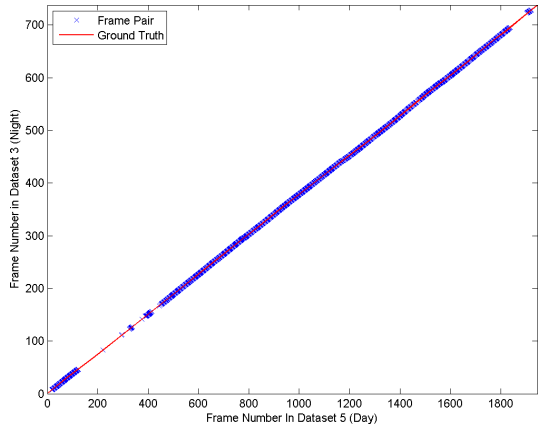

Fig. 8: Matched frame pairs from dataset 5 (day) to dataset 3 (night) for the Mt Cootha environment, overlaid on ground truth, for the fixed long exposure-variable short exposure datasets.

### D. Variable Short and Fixed Long Duration Exposures

Fig. 8 shows almost 80% of locations being matched with an average error of 0.62 frames. The period of false negatives after frame 100 was initiated by a sequence of overexposed images due to an approaching car's headlights. Fig. 7 shows the long and short exposure images from a matched image sequence.

### E. Frame Matching Analysis

One question these studies did not answer was whether it is possible to perform localization by matching individual, grayscale images without the key SeqSLAM processes of patch normalization and local neighborhood normalization. To answer this question, for every frame in dataset 5 we ranked the image matching scores produced by the image similarity calculation (Equation 1) with frames from dataset 3. We then identified where within that ranked list the correct image match (as determined by the ground truth data) was located and plotted the histograms shown in Fig. 9. With no extra processing, only 0.55% of the top matches as ranked by the image comparison algorithm were correct (Fig. 9a), with this fig. improving to 5.0% with both patch normalization and local neighborhood normalization. We concluded that relying on *individual* very low resolution images does not yield good localization performance.

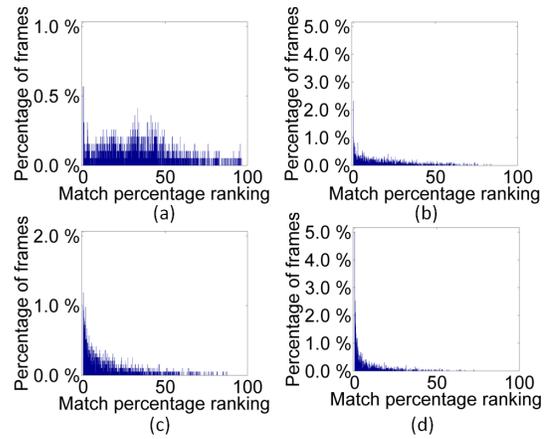

Fig. 9: At each time step, the individual image matching process ranks how closely all previous images match to the current image. This fig. shows a histogram of the ranking of the actual correct image match (as determined by ground truth) within the entire set of images, for image comparisons with (a) no extra processing (b) patch normalization (c) local neighborhood normalization and (d) both patch and local neighborhood normalization. Note the varied y-axis scales.

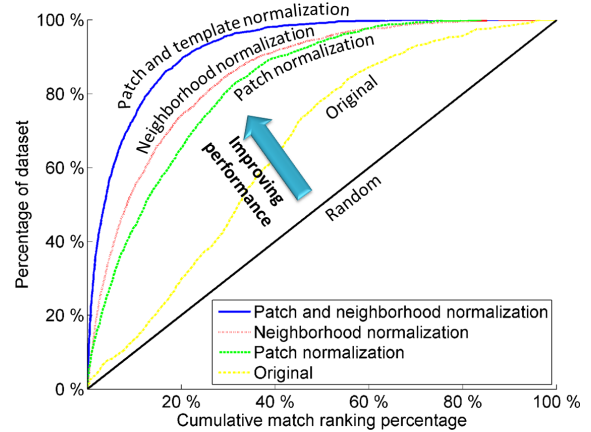

Fig. 10: Performing vanilla image matching matches the correct image matches only slightly better than at random chance. Introducing either patch or local neighborhood normalization results in a significant performance improvement, with both combined yielding the best performance.

To understand why matching using SeqSLAM works so much better than using unprocessed images, we produced the cumulative match ranking graph shown in Fig. 10. Without any extra processing, the correct image match is ranked barely better than at chance by the image comparison algorithm. However, the addition of patch normalization and local neighborhood normalization both massively improve the distribution, so that, while the correct image match is almost never ranked as the number one match candidate, it is always ranked very highly. 74% of the actual (as determined by ground truth) matching images are ranked in the top 10% of image match candidates, 89% are ranked in the top 20% of image matches and 99.2% of actual image matches are ranked in the top 50% of image matches. Local neighborhood normalization and to a lesser degree patch normalization vastly improve the average quality of image matches, making it easy for SeqSLAM to find coherent sequences of highly ranked (but usually not top ranked) image matches.

## VI. DISCUSSION AND FUTURE WORK

In this section we discuss the insights gained from this work as well as some of the practical implementation issues. Firstly, maximizing exposure duration and sensor gain is clearly unsuitable in applications where odometry information must be obtained from vision (as discussed in Section 2, current techniques can only extract motion information from simpler, smaller amounts of image blur than that dealt with in this paper). However, there are a large range of navigation applications where reasonable self-motion information is available, many car and wheeled robot applications being a major example. Future work will address opportunistic incorporation of self-motion information from wheel encoders or visual odometry.

If the method only worked with similar exposure durations, then achieving appropriate image exposure in bright sunlight and also in a dark night time environment with one camera would present a significant challenge. However, all four studies showed that matching is robust to significant discrepancies in exposure duration of more than a factor of 10. In fact, using more frequent, shorter exposure images during day-time appears to improve localization performance by providing a more fine-grained coverage of a route. This robustness to variation in exposure duration also removes the need to set the exposure duration based on vehicle velocity. A future area of investigation will be to trial logarithmic CMOS sensor types, which offer a much larger dynamic range. Additionally, the side effect of fixed-pattern noise on such sensors is likely to be less of a problem for the presented approach.

We also answered two questions: "what role does local neighborhood normalization and patch normalization play?" and "can you just use the original patch normalized images?" As shown in Fig. 10, performing straightforward image comparison on grayscale images leads to barely better than chance matching performance. Patch normalization improves performance by a significant margin, but it is the local neighborhood normalization that has the single biggest effect, massively improving the performance of the image matching algorithm. Future work will perform parameter studies to address the effect of varying the normalization neighborhood size and patch normalization variables.

## VII. CONCLUSION

The results presented in this paper show visual localization between bright day-time and dark night-time journeys through an environment is achievable using maximum exposure, maximum gain images at night, despite extreme image blur. Furthermore, higher frame rates (and hence shorter exposure durations) can be used during the day, without jeopardizing the ability to localize using long exposure night-time imagery. By demonstrating reliable localization using two relatively cheap consumer cameras, we hope to stimulate further research in vision-based processing at night. The low cost, compactness and passive sensing of cameras are great advantages, and it seems a shame to miss out on their benefits for half of every day.